\documentclass[runningheads]{llncs}

\usepackage{eccv}

\usepackage{eccvabbrv}

\usepackage{graphicx}
\graphicspath{{figures/}}
\usepackage{booktabs}

\usepackage[accsupp]{axessibility}  % Improves PDF readability for those with disabilities.

\usepackage{hyperref}

\usepackage{orcidlink}

\usepackage{wrapfig}

\begin{document}

\title{FPicker: Topology-Guided Evolution for Filament Tracing in Low-SNR Microscopy\thanks{Accepted to the 19th European Conference on Computer Vision (ECCV 2026).}}

\titlerunning{FPicker}

% Include the authors' ORCID for the camera-ready version, if at all possible.
\author{Tingyin Zhao\textsuperscript{1,2}\orcidlink{0009-0008-9111-8763} \and Mingtao Huang\textsuperscript{1,2,*}\orcidlink{0000-0002-8367-569X} \and
Yuan Shen\textsuperscript{1,2,*}\orcidlink{0000-0002-9396-1964}}

\authorrunning{T. Zhao et al.}

\institute{\textsuperscript{1}Department of Electronic Engineering, Tsinghua University, Beijing, China\\
\textsuperscript{2}Beijing National Research Center for Information Science and Technology, Beijing, China\\
\email{zhaoty25@mails.tsinghua.edu.cn, huangmt@mail.tsinghua.edu.cn, shenyuan\_ee@tsinghua.edu.cn}}

\maketitle
{\def\thefootnote{*}\footnotetext{Corresponding authors.}}

\begin{abstract}
	Automating filament tracing in Cryo-Electron Microscopy (Cryo-EM) is essential for 3D helical reconstruction but challenged by intersecting topologies and extremely low Signal-to-Noise Ratios ($\text{SNR} = \sigma_s^2/\sigma_n^2$ < 0.1 or -10 dB). Existing paradigms fail:  pixel-wise segmenters suffer from severe topological fracturing, box-based detectors face ghost center drift, sequential trackers derail due to error accumulation, and traditional active contours collapse under artificial closed-curve constraints. To resolve these bottlenecks, we present \textbf{FPicker}, the first topology-guided framework reconciling these incompatibilities. It unifies perception via a center-endpoint representation and an open-curve evolution module to explicitly model non-cyclic connectivity. On simulated benchmarks, FPicker outperforms top baselines by over $40\%$ relative gain in mean spatio-angular precision (mSAP) and reduces topological gap rates by over $60\%$ under extreme noise ($-20\text{ dB}$). By learning intrinsic physical geometry rather than local texture, FPicker demonstrates strong potential as a resilient geometric backbone. Its zero-shot performance on the real-world EMPIAR dataset exhibits robust topological resistance, achieving a state-of-the-art 82.9\% mSAP upon fine-tuning. Our results also suggest modeling physical priors is a highly robust path toward bridging the sim-to-real gap in signal-starved scientific imaging. The code is publicly available at: \url{https://github.com/tomzhaosky/FPicker}.
	
\end{abstract}

\section{Introduction}
\label{sec:intro}

Elucidating the atomic structure of filamentous proteins, such as amyloid fibrils involved in neurodegenerative diseases, is a cornerstone of modern structural biology and pharmaceutical science\cite{nogales2024bridging, scheres2023molecular, yang2022cryo}. Cryo-Electron Microscopy (Cryo-EM) has revolutionized this field; however, the initial step of filament picking---identifying and tracing individual instances from noisy micrographs, which explicitly enables accurate helical parameter estimation and subsequent 3D helical reconstruction \cite{huang2025accurate, he2017helical}---remains a laborious bottleneck. While automated picking for globular particles is well-established, filaments pose a unique topological challenge: they appear as flexible, continuous curvilinear structures that frequently run parallel, overlap, or intersect with one another \cite{vargas2024semantic, wu2021machine}. Tracing such complex structures is a fundamental problem that pervades biomedical vision, encompassing tasks like microtubule tracking in fluorescence microscopy and DNA strand tracing in nanoscale imaging\cite{masoudi2020instance, holmes2025quantifying, wagner2020two}. However, Cryo-EM represents an extreme case of this challenge because the filaments are embedded in vitreous ice, resulting in an extremely low Signal-to-Noise Ratio ($\text{SNR} = \sigma_s^2/\sigma_n^2$ < 0.1 or -10 dB) where the target signal is often indistinguishable from background noise at a local level.
\begin{figure}[t]
	\centering
	\includegraphics[width=\textwidth]{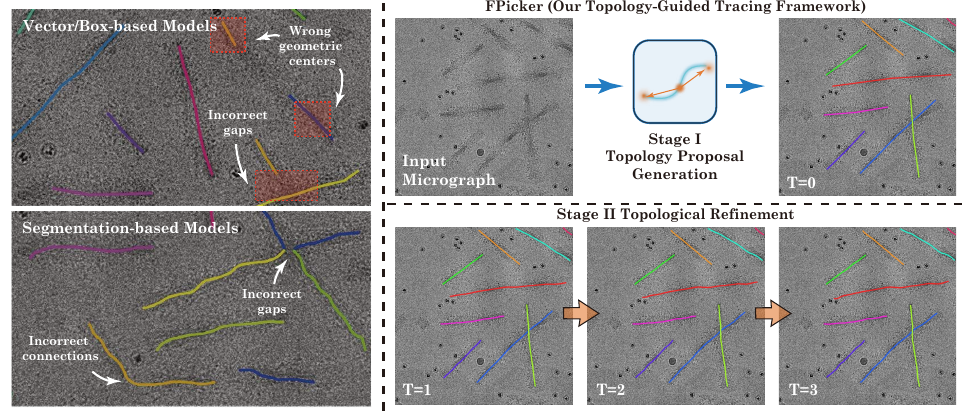}
	\caption{\textbf{Visualizing the topological gap and FPicker evolution.} Left: Geometric incompatibility of existing paradigms under extreme noise. Box-based detectors suffer from ghost centers (top), while segmentation models succumb to topological fracturing (bottom). Right: FPicker paradigm. Stage I (top) generates a linear topological prior via a center-endpoint representation. Stage II (bottom) iteratively refines this rigid prior ($T=1 \to 3$) via open-curve evolution. Please zoom in to see the details.}
	\label{fig:teaser}
\end{figure}

Existing vision paradigms face fundamental geometric incompatibilities here. Box-based detectors (\eg, crYOLO \cite{wagner2019sphire}, YOLOv8 \cite{yolov8_ultralytics}, etc.) trace filaments by linking sequential boxes. However, for thin flexible curves, a bounding box inherently encapsulates mostly background noise. This forces manual tuning of box sizes and causes the geometric centers to drift into signal-free regions (the ghost center degeneracy), leading to broken trajectories. Conversely, segmentation models (\eg, Topaz \cite{bepler2019positive}, SAM 2 \cite{ravi2024sam}, etc.) rely on a dense segment-then-skeletonize pipeline. While these models effectively extract intersecting contours, 2D projections force distinct filaments to share spatial pixels, creating ambiguities in disentangling individual instances and yielding merged topologies.

To explicitly model continuous topologies, methods turn to sequential tracking or active contours, yet both stumble in extreme noise. Sequential trackers\cite{liu2025netracer, shin2022deep} iteratively grow filaments but inherently derail into the signal-free background due to Markovian error accumulation. While active contour models (\eg, Deep Snake\cite{peng2020deep}, CurveGCN\cite{ling2019fast}, etc.) utilize Graph Convolutional Networks (GCNs) \cite{kipf2016semi} to holistically resist such local derailment, they suffer from two fatal flaws. First, standard box-based initializations force them to inherit the aforementioned ghost center problem, causing optimization collapse. Second, directly applying cyclic boundaries to open filaments creates an artificial tension pulling the tips together—a phenomenon we term shrinking degeneracy.

In this paper, we propose to \textit{break the closed loop}, as conceptually illustrated in Fig.~\ref{fig:teaser}. We introduce \textbf{FPicker} (\textbf{F}iber/\textbf{F}ilament \textbf{Picker}), a specialized framework designed for open-topology evolution in Cryo-EM. By replacing box priors with a center-endpoint representation and driving deformation via open-boundary graph convolutions, we ensure the model captures the continuous essence of protein filaments without succumbing to tip retraction. Our contributions are: 
\begin{itemize} 
	\item We introduce a center-endpoint representation to resolve the geometric ambiguity and ghost center problem inherent in box-based detection, effectively anchoring flexible filaments in extreme noise.
	\item We propose an open-curve evolution module to explicitly solve the fundamental endpoint shrinking limitation of traditional active contour models when applied to non-cyclic structures.
	\item We design a bridge curriculum strategy tailored for end-to-end low-SNR training, successfully decoupling the learning of local deformation physics from global localization variance.
\end{itemize}

Empirically, by optimizing within a strict physical forward model (Cryo-Sim), FPicker demonstrates strong potential as a resilient geometric backbone. It delivers effective zero-shot transfer and state-of-the-art fine-tuning performance on real-world EMPIAR micrographs. These results suggest that explicitly modeling intrinsic physical priors offers a fundamentally more robust path for AI for Science in signal-starved domains than brute-force data scaling.

\section{Related Works}
\label{sec:relatedworks}

\subsection{Vector and Box-based Detection}
Box-driven detectors (\eg, crYOLO, EPicker, YOLOv8/v10, RT-DETR, CenterNet, etc.)\cite{yolov8_ultralytics, wagner2019sphire, zhang2022epicker, wang2024yolov10, zhou2019objects, zhao2024detrs} excel at globular particles. To trace continuous filaments, they typically detect local segments and link their box centers. We identify a geometric incompatibility here: representing 1D flexible curves with 2D static bounding boxes is intrinsically ill-posed. Because a box inherently encapsulates mostly background noise rather than the thin filament itself, it necessitates tedious manual tuning of anchor sizes. More critically, the geometric center of these boxes frequently falls into the signal-free background (the ghost center degeneracy), derailing accurate linking and feature extraction. FPicker departs from this paradigm by anchoring directly onto the topological backbone.

\subsection{Segmentation-based Approaches}
The segment-then-skeletonize paradigm, widely adopted in general vision and tubular tracing (\eg, Topaz, U-Net, SAM 2, DeepVesselNet, etc.) \cite{bepler2019positive, ma2024u, ma2024segment, ronneberger2015u, xie2021segformer, ravi2024sam, tetteh2020deepvesselnet, bastani2018roadtracer}, avoids box constraints. While effective at extracting overall filament contours, these methods struggle to distinguish individual instances. Because Cryo-EM micrographs are inherently 2D projections, intersecting filaments share identical spatial pixels. Consequently, even when advanced continuity constraints are applied to mitigate fracturing, grid-bound representations inevitably merge distinct instances at 2D crossovers. Furthermore, non-differentiable skeletonization decouples topology from learning, preventing the network from utilizing global connectivity to heal noise-induced gaps or resolve junctions.

\subsection{Explicit Topological Tracking and Evolution}
Methods explicitly modeling topology fall into two paradigms: sequential tracking and active contours. Sequential trackers\cite{liu2025netracer, shin2022deep, dai2019deep, liu2016rivulet}, including those utilizing Bayesian probabilistic inference \cite{huang2025bayesian}, iteratively extend trajectories from local seeds. However, under extreme Cryo-EM noise, misguided local gradients cause trackers to irreversibly derail into the background due to Markovian error accumulation. Conversely, deep active contours (\eg, Deep Snake, CurveGCN, BoundaryFormer, DSAC, RLS, etc.)\cite{peng2020deep, ling2019fast, lazarow2022instance, marcos2018learning, le2018reformulating} holistically deform explicit graphs, naturally resisting local derailment. Yet, directly adapting them to open filaments faces two fatal bottlenecks: an initialization trap (inheriting box-based ghost centers) and a manifold mismatch (cyclic convolutions artificially shrink open endpoints together). Ultimately, FPicker synergizes the open-topology essence of sequential tracking with the holistic deformation of active contours, resolving their respective bottlenecks via a center-endpoint prior and strictly open-boundary evolution.

\section{Methods}
\label{sec:methods}
\subsection{Overview}
\begin{figure}[h]
	\centering
	\includegraphics[width=\textwidth]{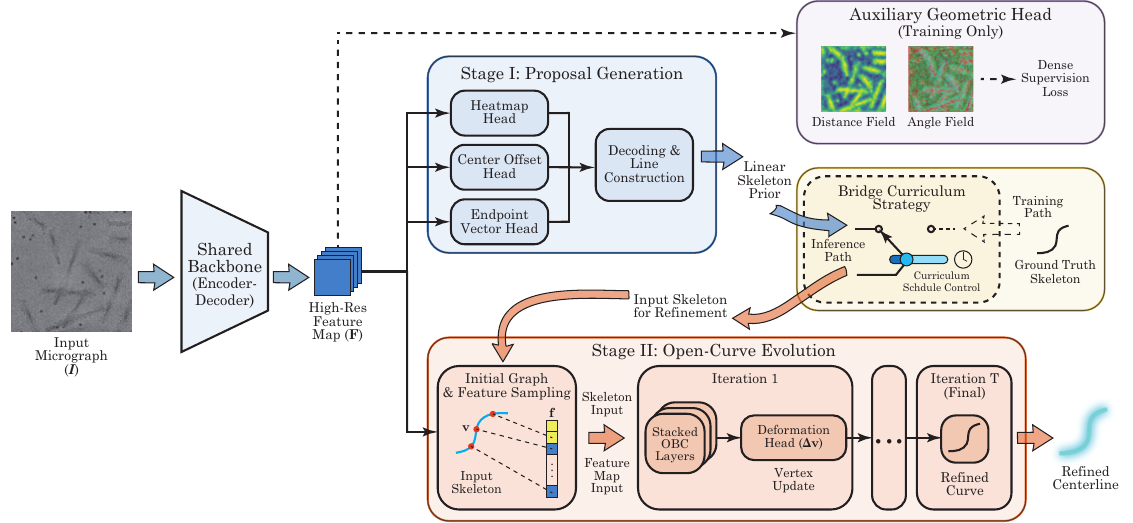}
	\caption{\textbf{FPicker Architecture.} The shared backbone extracts a high-resolution feature map $\mathbf{F}$ to drive a two-stage pipeline. Stage I generates linear skeletal priors using a center-endpoint representation. Stage II iteratively refines these priors via an open-curve snake module. During training, a bridge curriculum strategy (dashed box) and an auxiliary geometric head ensure stable convergence in extreme noise conditions.}
	\label{fig:architecture}
\end{figure}
The overall architecture of FPicker is illustrated in Fig.~\ref{fig:architecture}. We formulate filament tracing as a unified, end-to-end trainable framework that maps a raw Cryo-EM micrograph $\mathcal{I}$ to a set of precise, open skeletal trajectories $\mathcal{F}$. The pipeline follows a coarse-to-fine paradigm, consisting of a shared feature extractor and two specialized stages, which are all gradient accessible:
\begin{enumerate}
	\item \textbf{Shared Feature Extraction:} A deep encoder-decoder backbone \cite{he2016deep, yu2018deep, liu2021swin} processes the micrograph $\mathcal{I}$ into a high-resolution semantic feature map $\mathbf{F} \in \mathbb{R}^{H/s \times W/s \times C}$ (stride $s$). $\mathbf{F}$ serves as the unified representation for detection and vertex-level refinement.
	
	\item \textbf{Proposal Generation (Stage I):} Operating on  the feature map $\mathbf{F}$, a multi-head branch predicts dense topological heatmaps and center-endpoint fields, aggregating them into linear skeletal priors $\hat{\mathcal{S}}_{\text{init}}$ to anchor instances despite extreme noise.
	
	\item \textbf{Topological Refinement (Stage II):} The linear skeletons serve as the initial state for the open-curve evolution module. For each instance, vertex-specific features are bilinearly sampled from $\mathbf{F}$. A graph convolutional network (GCN) then iteratively deforms the rigid priors into flexible curves that snap to the true protein skeletal trajectories.
\end{enumerate}

To bridge the optimization gap between these stages in low-SNR regimes, we introduce a bridge curriculum strategy and auxiliary geometric fields to enforce structural consistency during the training process.

\subsection{Topology-Aware Proposal Generation}
To enforce structural alignment, we redefine the target as a topological centroid $\mathbf{c}_i$, strictly constrained to the filament trajectory $\mathbf{\mathcal{G}}_i(\ell)$ at the arc-length median: $\mathbf{c}_i = \mathcal{G}_i(L_i/2)$. This ensures the detector anchors onto discriminative protein density rather than background artifacts. 

To extract discrete instances, we apply a $3 \times 3$ max-pooling operation over the topological heatmaps for non-maximum suppression (NMS) \cite{zhou2019objects}. Crucially, by operating directly on the smoothed semantic feature space, this topological peak extraction is highly stable and completely circumvents the need for bounding box regression. Local peaks form discrete valid centroids $\mathbf{c}_i$.

Complementing this, we regress dense displacement fields $\{\hat{\mathbf{d}}_{\text{start}}, \hat{\mathbf{d}}_{\text{end}}\}$ from these centers. By uniformly interpolating $N$ vertices between $\mathbf{c}_i + \hat{\mathbf{d}}_{\text{start}}$ and $\mathbf{c}_i + \hat{\mathbf{d}}_{\text{end}}$, we establish the linear topological prior $\hat{\mathcal{S}}_{\text{init}}$, which acts as a geometric attractor, defining a rotation-aware capture range that ensures even high-curvature sinusoidal fibrils fall within the subsequent GCN's receptive field.

\subsection{Open-Curve Evolution}
While the linear skeleton $\hat{\mathcal{S}}_{\text{init}}$ provides a coarse localization, biological filaments possess non-linear elasticity and variable curvatures. To capture these fine-grained geometries, we adapt the active contour framework into a topology-specific open-curve evolution module. Unlike segmentation-based refinement, our approach treats the filament as a continuous, deformable entity, ensuring axial integrity even in fragmented signal regions.

\textbf{Graph Construction and Feature Sampling.}
\begin{wrapfigure}{R}{0.38\textwidth}
	\centering
	\includegraphics[width=\linewidth]{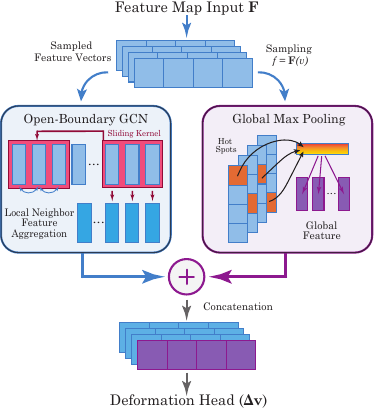} 
	\caption{\textbf{Dual-Stream Architecture.} Local (OBC) and Global feature streams fuse to guide deformation.}
	\label{fig:open_curve_arch}
\end{wrapfigure}
Given a predicted linear skeleton $\hat{\mathcal{S}}_{\text{init}}$ generated from the topological centroid $\hat{\mathbf{c}}$ and displacement vectors $\{\hat{\mathbf{d}}_{\text{start}}, \hat{\mathbf{d}}_{\text{end}}\}$, we uniformly sample $N$ vertices $\hat{\mathcal{V}} = \{\hat{\mathbf{v}}_i\}_{i=1}^N$ along the axis. Unlike traditional closed-polygon snakes, we construct an open graph connecting each internal vertex $\hat{\mathbf{v}}_i$ ($1 < i < N$) only to its immediate neighbors.

For each $\hat{\mathbf{v}}_i = (\hat{x}_i, \hat{y}_i)$ sampled along the initial skeleton, we extract a feature vector $\mathbf{f}_i$ from $\mathbf{F}$ via bilinear interpolation: $\mathbf{f}_i = \text{Interpolate}(\mathbf{F}, \hat{\mathbf{v}}_i)$. Interpolating directly along the skeleton yields a deformable receptive field precisely tailored to the true trajectory. This feature-centric approach avoids explicit geometric embeddings prone to noise-induced drift and bypasses the ghost center vulnerability inherent in box-based RoI pooling.

\textbf{Open-Boundary Graph Convolution (OBC).}
Standard active contour implementations employ circular convolution, enforcing periodic boundary conditions where vertex $v_N$ connects to $v_1$. This introduces a phantom tension for open filaments, causing tip retraction. To achieve open-curve evolution for continuous filaments, we reformulate the traditional cyclic GCNs by strictly imposing open-boundary conditions during feature aggregation.

Mathematically, we employ replicate padding at the graph boundaries. For a graph signal $\mathbf{f} \in \mathbb{R}^N$, the feature aggregation for a vertex $i$ is constrained to its geodesic neighbors:
\begin{equation}
	\mathbf{f}'_i = \sum_{j=-d}^{d} \mathbf{W}_j \cdot \mathbf{f}_{\max(1, \min(N, i+j))}
\end{equation}
This truncation breaks the gradient flow between logically distant endpoints, setting the tension at the tips to zero and preventing the shrinking degeneracy.

\textbf{Iterative Deformation with Global Structural Anchoring.}
In extreme noise, local receptive fields are often dominated by stochastic background fluctuations, causing individual vertices to drift. To ensure structural consistency without relying on fragile coordinate assumptions, we introduce a global structural anchoring mechanism.

At each iteration $t$, vertex features $\mathbf{F}^{(t)}$ are processed through two parallel streams, as illustrated in Fig.~\ref{fig:open_curve_arch}. The local stream extracts geometric affinities via stacked OBC layers, while the global stream employs a max-pooling operation to capture the most salient signal. Crucially, because the shared feature $\mathbf{F}$ is explicitly regularized by auxiliary low-pass geometric fields in Section~\ref{subsec:aux_geometric_field}, this pooling operation naturally resists high-frequency noise disturbance:
\begin{equation}
	\mathbf{h}_i^{\text{local}} = \Phi_{\text{GCN}}(\mathbf{f}_i^{(t)}), \quad \mathbf{h}^{\text{global}} = \max_{j=1}^N \{ \Phi_{\text{init}}(\mathbf{f}_j^{(t)}) \}
\end{equation}
We posit that in low-SNR regimes, data-dependent priors outperform independent ones. Traditional positional embeddings impose rigid absolute coordinates that mislead when the initial snake drifts due to noise. In contrast, our $\mathbf{h}^{\text{global}}$ is data-dependent: it dynamically aggregates the strongest filament existence signals (e.g., from a clearly visible segment) regardless of absolute position.

This global descriptor acts as a semantic anchor, broadcasting high-confidence structural cues to vertices in low-signal regions. The deformation offset is then predicted by applying a learnable weight matrix $\mathbf{W}_{\text{offset}}$ and bias $\mathbf{b}_{\text{offset}}$ to the concatenated features:
\begin{equation}
	\hat{\mathbf{v}}_i^{(t+1)} = \hat{\mathbf{v}}_i^{(t)} + \Delta \hat{\mathbf{v}}_i^{(t)}, \quad \Delta \hat{\mathbf{v}}_i^{(t)} = \mathbf{W}_{\text{offset}} {\begin{bmatrix} \mathbf{h}_i^{\text{local}} \ \mathbf{h}^{\text{global}} \end{bmatrix}}^{\top} + \mathbf{b}_{\text{offset}}
\end{equation}
By conditioning deformation on this global context, the network effectively hallucinates the correct trajectory for noise-submerged vertices, preventing the fragmentation typical of local-only methods.

\begin{figure*}[h]
	\centering
	\includegraphics[width=\textwidth]{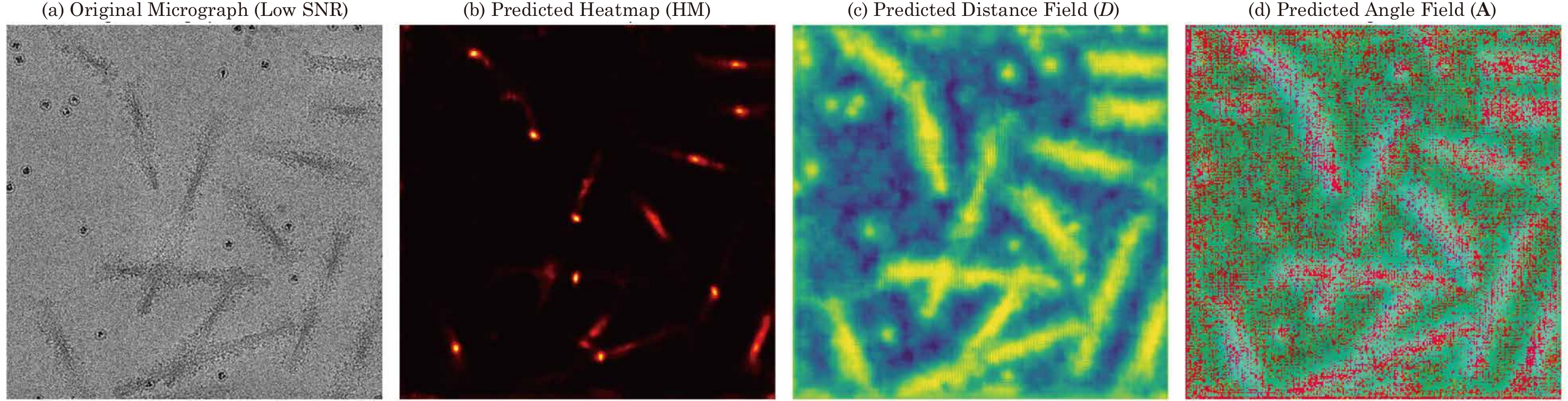}
	\caption{\textbf{Visualization of Auxiliary Fields.} (a) Raw input micrograph. (b) Predicted topological centroid heatmap. (c) Distance Field ($\hat{\mathcal{D}}$), revealing smooth tubular manifolds. (d) Angle Field ($\hat{A}$) in HSV space (Hue: tangent direction; Value: distance). The continuous color gradients demonstrate robust learning of intrinsic filament geometry.}
	\label{fig:aux_fields}
\end{figure*}

\subsection{Auxiliary Geometric Field Learning.}
\label{subsec:aux_geometric_field}
While the evolution module effectively captures the filament's elasticity, its convergence heavily relies on the quality of the underlying feature map. In extreme low-SNR regimes, intensity-based features are prone to local minima. To regularize the backbone and enforce structural awareness, we attach an auxiliary geometric branch to the backbone to regress two dense geometric fields: a Distance Field ($\mathcal{D}$) representing Euclidean proximity to the nearest trajectory, and a 2-channel Angle Field ($\mathbf{A}$) encoding local tangent directions $(\cos \theta, \sin \theta)$.

Inspired by DeepLSD \cite{pautrat2023deeplsd}, our dual-field representation establishes robust dense geometric supervision. While conceptually resembling Local Shape Descriptors (LSDs) \cite{sheridan2023local}, FPicker departs from their algorithmic role: rather than strengthening local pixel classifications, our fields act as continuous low-pass regularizers guiding sparse center-endpoint proposals and holistic open-curve graph evolution, as visually demonstrated in Fig.~\ref{fig:aux_fields}. Consequently, unless supported by a coherent center-endpoint prior, the impact of spurious auxiliary-field artifacts is mitigated, rendering our framework resilient to Contrast Transfer Function (CTF) oscillations inherent in Cryo-EM defocus imaging.

\subsection{The Bridge Curriculum Strategy}
End-to-end training in extreme noise faces a cold start dilemma: the evolution module requires stable initialization to learn topology, yet the early-stage detector yields chaotic proposals. To prevent optimization collapse, we propose the bridge curriculum strategy, designed to decouple the learning of local deformation physics from global localization variance. We employ a transition regulator $P_{\text{gt}}$ to modulate the input $\mathcal{S}_{\text{input}}$:
\begin{equation}
	\mathcal{S}_{\text{input}} = \beta (\mathcal{S}_{\text{gt}} + \epsilon) + (1 - \beta) \hat{\mathcal{S}}_{\text{pred}}, \quad \text{where} \quad \beta \sim \text{Bernoulli}(P_{\text{gt}})
\end{equation}
where $\epsilon$ denotes Gaussian perturbation. Prior to this element-wise modulation, both the ground-truth skeleton $\mathcal{S}_{\text{gt}}$ and the predicted linear prior $\hat{\mathcal{S}}_{\text{pred}}$ are uniformly resampled to exactly $N$ equidistant vertices (e.g., $N=128$) to guarantee strict dimensional alignment. Unlike standard teacher forcing, this stochastic injection forces the evolution module to first master the physical laws of feature attraction (i.e., snapping to ridge intensity) within a controlled capture range ($P_{\text{gt}}=1$ for the first 10 epochs), independent of detection drift. As the detector stabilizes, we linearly anneal $P_{\text{gt}} \to 0$ over the subsequent 110 epochs, gradually exposing the refinement module to the inherent variance of predicted priors until fully autonomous inference is achieved.

\subsection{Loss Function}
The FPicker framework is trained via a multi-task objective function. The total loss $\mathcal{L}_{total}$ is defined as a weighted sum of detection, geometric, and evolution components, ensuring the model balances coarse localization with fine-grained structural alignment:
\begin{equation}
	\mathcal{L}_\text{total} = \lambda_\text{hm}\mathcal{L}_\text{hm} + \lambda_\text{reg}\mathcal{L}_\text{reg} + \lambda_\text{ends}\mathcal{L}_\text{ends} + \lambda_\text{aux}\mathcal{L}_\text{aux} + \lambda_\text{evol}\mathcal{L}_\text{evol}
\end{equation}
where $\lambda$ are coefficients used to balance the contribution of each task.

\textbf{Topology-Anchored Heatmap Loss ($\mathcal{L}_\text{hm}$).} To handle extreme foreground-background imbalance, we employ a modified Focal Loss\cite{lin2017focal}. Inspired by the objective formulation of CenterNet\cite{zhou2019objects}, our loss is tailored to anchor strictly onto the topological centroids $\mathbf{c}_i = \mathcal{G}_i(L_i/2)$ rather than ill-posed bounding box centers. Let $\hat{Y}(\mathbf{p})$ and $Y(\mathbf{p})$ denote the predicted and ground-truth Gaussian maps at spatial location $\mathbf{p} \in \Omega$. The loss is defined as:
\begin{equation}
	\mathcal{L}_\text{hm} = -\frac{1}{N} \sum_{\mathbf{p} \in \Omega}
	\begin{cases}
		(1 - \hat{Y}(\mathbf{p}))^\alpha \log(\hat{Y}(\mathbf{p})) & \text{if } Y(\mathbf{p})=1 \\
		(1 - Y(\mathbf{p}))^\beta (\hat{Y}(\mathbf{p}))^\alpha \log(1 - \hat{Y}(\mathbf{p})) & \text{otherwise}
	\end{cases}
\end{equation}
where $N$ is the number of valid filaments, and $\alpha=2, \beta=4$ are hyper-parameters controlling the penalty landscape around the topological medians.

\textbf{Regression Losses ($\mathcal{L}_\text{reg}$ and $\mathcal{L}_\text{ends}$).} The center offset map $\hat{\mathbf{o}}$ and endpoint vector map $\hat{\mathbf{d}}$ are trained using L1 loss. Crucially, to avoid noise interference, we apply a sparse supervision strategy: the loss is computed only at the ground-truth center locations $\mathbf{c}_k$.
\begin{equation}
	\mathcal{L}_{\text{ends}} = \frac{1}{N} \sum_{k=1}^{N} \left \| \hat{\mathbf{d}}(\mathbf{c}_k) - \mathbf{d}_k \right \|_1, 
	\quad \mathcal{L}_{\text{reg}} = \frac{1}{N} \sum_{k=1}^{N} \left \| \hat{\mathbf{o}}(\mathbf{c}_k) - \left( \frac{\mathbf{c}_k}{s} - \lfloor \frac{\mathbf{c}_k}{s} \rfloor \right) \right \|_1 
\end{equation}
where $s$ is the output stride. This forces the network to focus its capacity on valid structural instances.

\textbf{Auxiliary Geometric Loss ($\mathcal{L}_\text{aux}$).} Unlike the sparse regression heads, the auxiliary fields provide dense supervision. We compute the L1 difference between the predicted fields ($\hat{\mathcal{D}}, \hat{\mathbf{A}}$) and the ground truth ($\mathcal{D}, \mathbf{A}$) over the entire valid fiber mask $M$:
\begin{equation}
	\mathcal{L}_{\text{aux}} = \frac{1}{|M|} \sum_{(x,y) \in M} \left( |\hat{\mathcal{D}}_{xy} - \mathcal{D}_{xy}| + \| \hat{\mathbf{A}}_{xy} - \mathbf{A}_{xy} \|_2 \right)
\end{equation}
This term acts as a regularizer, preventing the backbone from overfitting to background noise patterns.

\textbf{Open-Curve Evolution Loss ($\mathcal{L}_\text{evol}$).} 
To capture the flexible geometry of filaments while maintaining structural integrity, we define the topological refinement loss $\mathcal{L}_\text{evol} = \mathcal{L}_\text{fit} + \lambda_\text{uni}\mathcal{L}_\text{uni}$. Here, $\mathcal{L}_\text{fit}$ employs the Smooth-L1 criterion to minimize the discrepancy between evolved vertices $\hat{\mathcal{V}}$ and ground-truth points $\mathcal{V}$, while $\mathcal{L}_\text{uni}$ penalizes edge-length variance to prevent vertex clustering in noise-dominated regions:
\begin{equation}
	\mathcal{L}_\text{fit} = \frac{1}{N} \sum_{i=1}^N \text{Smooth}_\text{L1}(\hat{\mathbf{v}}_i, \mathbf{v}_i), \quad \mathcal{L}_\text{uni} = \frac{1}{N-1} \sum_{i=1}^{N-1} \left| \|\hat{\mathbf{v}}_{i+1} - \hat{\mathbf{v}}_i\|_2 - \bar{d} \right|
\end{equation}
where $\bar{d}$ denotes the average predicted edge length. This combined objective ensures that the open curve not only accurately snaps to the true skeleton topology but also maintains a physically plausible, equidistant vertex distribution, which is critical for bridging fragmented signals in extreme low-SNR environments.

\section{Experiments}
\label{sec:exps}

\subsection{Experimental Setup}

\textbf{Datasets and Benchmarks.} Lacking open-source datasets with pixel-level filament annotations, we eschew large-scale supervised training on real data. We utilize: \textit{(1) Cryo-Sim:} An electron optics simulation engine generating 20,000 micrographs (explicitly modeling CTF physics like spherical aberration, amplitude contrast, and B-factor decay; full details in the Supplementary Material). It contains micrographs across High (SNR 0.1, -10 dB), Medium (0.05, -13 dB), and Extreme (0.01, -20 dB) noise regimes. \textit{(2) Custom-EMPIAR:} A real-world dataset of 500 manually annotated micrographs from EMPIAR-10230 and EMPIAR-10340 \cite{iudin2016empiar, zhang2020novel, falcon2018tau}, containing $\sim$5,500 individual filaments. Models are trained/tested on Cryo-Sim, evaluated zero-shot on EMPIAR, and finally fine-tuned on EMPIAR. Both datasets can be found in open-sourced materials. 

\textbf{Implementation \& Metrics.} FPicker is implemented in PyTorch and trained on 5 RTX 4090 GPUs using Adam (LR $1e-4$) for 200 epochs. To comprehensively assess performance, we first establish mSAP (mean spatio-angular precision) as our core matching criterion, evaluating spatial Chamfer distance \cite{barrow1977parametric} under a strict angular tolerance ($\Delta\theta < 15^\circ$). Based on mSAP true positives, we compute standard detection metrics: F1-score and gap rate ($1 - \text{mAR}$). For structural continuity, we evaluate standard clDice \cite{shit2021cldice} alongside our proposed fp-clDice, which incorporates an exponential fragmentation penalty ($\exp(-0.1 (N_{\text{frag}} - 1))$) to heavily penalize dashed predictions. Finally, we report P-Ang (penalized angle error), assigning a $90^\circ$ penalty to unrecovered instances to prevent selection bias. Preliminary variance analysis of FPicker performance is also included in the supplementary material.

\subsection{Comparison with State-of-the-Arts}
\begin{figure*}[h]
	\centering
	\includegraphics[width=\textwidth]{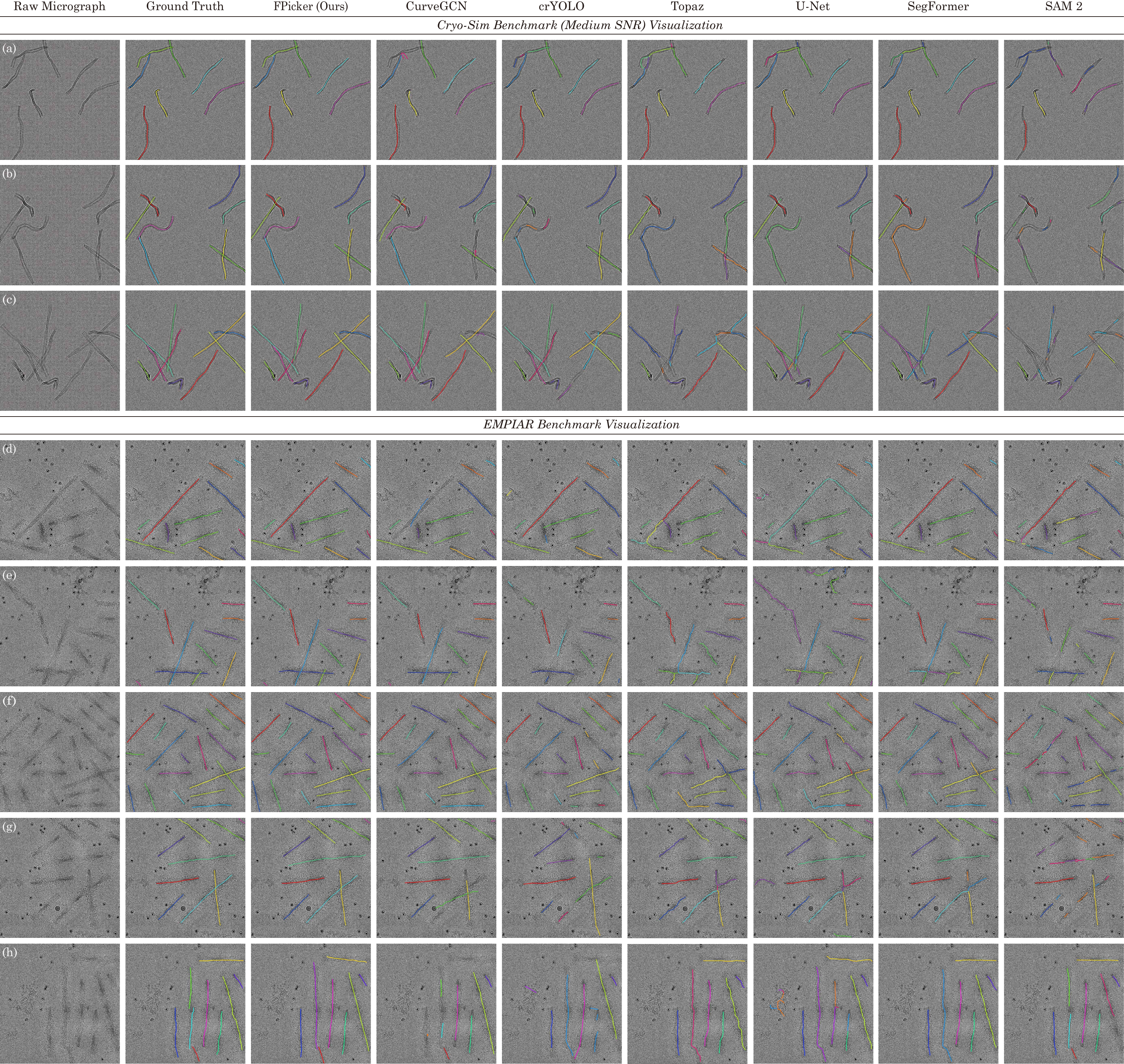}
	\caption{\textbf{Qualitative Comparison.} Box detectors show severe discontinuities, while segmenters merge instances at crossovers. Closed contours perform poorly at endpoints, whereas FPicker disentangles junctions and recovers continuous skeletons under extreme noise. Colors distinguish instances. Please zoom in to see the details.}
	\label{fig:qualitative_comparison}
\end{figure*}

We benchmark FPicker against three dominant paradigms: Box-based Detection (crYOLO, YOLOv8), Pixel-wise Segmentation (Topaz, U-Net, SegFormer, SAM 2), and Closed Active Contours (Deep Snake, CurveGCN). While Table~\ref{tab:snr_comparison} reports the quantitative performance across different noise regimes, Fig.~\ref{fig:qualitative_comparison} provides the corresponding qualitative results that visually demonstrate FPicker's robustness under extreme noise.

\textbf{Ghost Center Collapse (Box-based Methods).}
In High SNR regimes, box-based methods like YOLOv8 and crYOLO perform robustly (mSAP 78.5\% and 75.2\%). However, performance degrades drastically under Extreme noise (-20 dB), with YOLOv8 and crYOLO plummeting to \textbf{28.5\%} and \textbf{25.4\%} mSAP, respectively. This confirms our hypothesis regarding the ghost center problem: for thin, curved filaments buried in noise, the geometric center of a bounding box often resides in the background signal. Without a strong visual feature at the anchor point, the detector struggles to converge, resulting in severe recall failure (e.g., YOLOv8 Gap Rate \textbf{45.2\%}).

\textbf{Topological Fracturing (Segmentation Methods).}
While U-Net and SegFormer maintain better localization than box detectors at Medium SNR (mSAP $\sim$62-64\%), they suffer from severe topological inconsistency. This is quantified by their high Gap Rates (\textbf{18.5\%} and \textbf{16.2\%} respectively) compared to FPicker (6.3\%). The pixel-wise independence assumption causes these models to treat noise fluctuations as boundaries, resulting in dashed line predictions that require heavy post-processing to repair. Despite using a skeletal outline slightly larger than the ground truth---an exceptionally strong prompt bordering on explicit localization---SAM 2 yields an mSAP of only \textbf{12.1\%} at Extreme SNR. SAM 2 fundamentally fails because its attention mechanism relies on high-frequency, texture-based affinities, which systematically collapse in the textureless, shot-noise-dominated Cryo-EM domain.

\textbf{The Shrinking Degeneracy (Closed Active Contours).}
Deep Snake effectively mitigates fracturing (Gap Rate 12.5\% at Medium SNR), validating the benefit of topological modeling. However, it fails geometrically. Its Penalized Angle Error is significantly high (\textbf{26.2$^\circ$} at Medium SNR) compared to FPicker (\textbf{15.5$^\circ$}). This is the direct consequence of the endpoint shrinking problem: the cyclic convolution forces the open filament tips to pull towards each other, distorting the tangent direction at the extremities.

\textbf{FPicker Superiority.}
FPicker (DLA-34) achieves decisive superiority, particularly in the Extreme SNR regime. It maintains a low Gap Rate (9.8\%) and acceptable Angle Error (20.5$^\circ$), proving that the open-curve evolution successfully decouples internal smoothness from endpoint tension.

\begin{table}[t]
	\caption{\textbf{Main Comparison across SNR regimes on Cryo-Sim.} Comparisons are grouped by paradigm. \textbf{F1}: F1-Score ($\% \uparrow$), \textbf{clD}: clDice ($\% \uparrow$), \textbf{mSAP}: Spatio-Angular Precision ($\% \uparrow$), \textbf{fp-clD}: Fragmentation-Penalized clDice ($\% \uparrow$), \textbf{Gap}: Gap Rate ($\% \downarrow$), \textbf{P-Ang}: Penalized Tangent Error ($^\circ \downarrow$).}
	\label{tab:snr_comparison}
	\centering
	\setlength{\tabcolsep}{1.6pt}
	\resizebox{\textwidth}{!}{
		\begin{tabular}{@{}l|cccccc|cccccc|cccccc@{}}
			\toprule
			& \multicolumn{6}{c|}{\textbf{High SNR (0.1/-10 dB)}} & \multicolumn{6}{c|}{\textbf{Medium SNR (0.05/-13 dB)}} & \multicolumn{6}{c}{\textbf{Extreme SNR (0.01/-20 dB)}} \\
			Method & 
			\makebox[3em][c]{F1$\uparrow$} & \makebox[3em][c]{clD$\uparrow$} & \makebox[3.2em][c]{mSAP$\uparrow$} & \makebox[3.2em][c]{fp-clD$\uparrow$} & \makebox[3em][c]{Gap$\downarrow$} & \makebox[3em][c]{P-Ang$\downarrow$} & 
			\makebox[3em][c]{F1$\uparrow$} & \makebox[3em][c]{clD$\uparrow$} & \makebox[3.2em][c]{mSAP$\uparrow$} & \makebox[3.2em][c]{fp-clD$\uparrow$} & \makebox[3em][c]{Gap$\downarrow$} & \makebox[3em][c]{P-Ang$\downarrow$} & 
			\makebox[3em][c]{F1$\uparrow$} & \makebox[3em][c]{clD$\uparrow$} & \makebox[3.2em][c]{mSAP$\uparrow$} & \makebox[3.2em][c]{fp-clD$\uparrow$} & \makebox[3em][c]{Gap$\downarrow$} & \makebox[3em][c]{P-Ang$\downarrow$} \\
			\midrule
			\multicolumn{19}{l}{\textit{Box/Vector-based Models}} \\
			crYOLO\cite{wagner2019sphire} & 79.5 & 86.8 & 75.2 & 85.4 & 6.8 & 14.2 & 55.4 & 64.1 & 48.6 & 52.1 & 24.5 & 48.5 & 35.2 & 44.1 & 25.4 & 31.5 & 48.2 & 64.4 \\
			YOLOv8\cite{yolov8_ultralytics} & 82.0 & 89.2 & 78.5 & 88.1 & 5.1 & 12.5 & 58.5 & 68.2 & 52.4 & 56.5 & 20.2 & 45.1 & 38.5 & 48.2 & 28.5 & 35.4 & 45.2 & 62.1 \\
			\midrule
			\multicolumn{19}{l}{\textit{Segmentation-based Models\textsuperscript{\dag}}} \\
			Topaz\cite{bepler2019positive} & 78.5 & 86.5 & 72.1 & 85.4 & 6.8 & 14.8 & 54.2 & 64.5 & 48.1 & 52.6 & 24.5 & 49.8 & 34.6 & 44.5 & 24.2 & 31.8 & 48.5 & 65.4 \\
			U-Net\cite{ronneberger2015u} & 89.1 & 92.5 & 86.4 & 91.2 & 2.4 & 8.2 & 68.5 & 75.4 & 62.1 & 61.2 & 18.5 & 32.4 & 46.5 & 55.4 & 38.2 & 42.1 & 38.5 & 55.2 \\
			SegFormer\cite{xie2021segformer} & 89.8 & 93.1 & 87.1 & 91.8 & 2.1 & 7.9 & 71.2 & 77.1 & 64.8 & 63.5 & 16.2 & 29.8 & 49.2 & 58.1 & 41.5 & 45.3 & 35.4 & 52.8 \\
			SAM 2\cite{ravi2024sam} & 60.1 & 68.5 & 65.2 & 55.2 & 45.0 & 22.1 & 42.1 & 48.5 & 35.4 & 28.2 & 52.6 & 55.4 & 18.5 & 25.4 & 12.1 & 15.2 & 75.2 & 82.5 \\
			\midrule
			\multicolumn{19}{l}{\textit{Active Contours}} \\
			Deep Snake\cite{peng2020deep} & 90.5 & 93.5 & 88.2 & 92.5 & 1.5 & 9.8 & 75.4 & 81.5 & 68.2 & 71.4 & 12.5 & 26.2 & 54.5 & 62.4 & 45.8 & 51.2 & 28.5 & 45.2 \\
			CurveGCN\cite{ling2019fast} & 90.2 & 92.8 & 88.5 & 91.5 & 1.8 & 10.5 & 76.8 & 82.1 & 71.5 & 73.2 & 11.8 & 24.5 & 56.8 & 64.1 & 48.2 & 53.5 & 26.2 & 42.8 \\
			\midrule
			\multicolumn{19}{l}{\textbf{Ours (Ablation)}} \\
			FPicker (ResNet-50) & 92.5 & 94.2 & 89.8 & 93.5 & 1.5 & 2.1 & 87.2 & 88.8 & 89.0 & 84.5 & 7.1 & 19.9 & 75.4 & 81.2 & 72.1 & 72.5 & 12.2 & 24.5 \\
			FPicker (Swin-T) & 93.8 & 95.5 & 91.5 & 94.8 & 0.8 & 1.6 & 88.5 & 89.4 & 90.5 & 83.9 & 7.3 & 19.4 & 78.2 & 83.5 & 74.8 & 75.8 & 10.5 & 21.8 \\
			\textbf{FPicker (DLA-34)} & \textbf{94.5} & \textbf{96.1} & \textbf{92.1} & \textbf{95.8} & \textbf{0.7} & \textbf{1.4} & \textbf{89.6} & \textbf{90.7} & \textbf{91.1} & \textbf{89.6} & \textbf{6.3} & \textbf{15.5} & \textbf{80.1} & \textbf{85.2} & \textbf{76.5} & \textbf{78.5} & \textbf{9.8} & \textbf{20.5} \\
			\bottomrule
		\end{tabular}
	}
	
	\raggedright
	\scriptsize{\textsuperscript{\dag} \textbf{Note:} Segmentation maps are skeletonized using standard medial axis algorithms, which has been widely applied in medical image analysis.}
\end{table}

\subsection{Sim-to-Real Transfer on EMPIAR}
To evaluate real-world applicability, we benchmark the Sim-to-Real transfer capability on the Custom-EMPIAR dataset (Table~\ref{tab:empiar_transfer}). Recognizing the extreme domain gap, we evaluate all models in two phases: purely zero-shot (trained only on Cryo-Sim) and fine-tuned (adapted on EMPIAR).
\begin{table}[h]
	\centering
	\begin{minipage}[c]{0.64\textwidth}
		\centering
		\caption{Transfer Performance of FPicker (DLA-34) on the Custom-EMPIAR Real-World Dataset. Format: \textit{Zero-Shot / Fine-Tuned}}
		\label{tab:empiar_transfer}
		\renewcommand{\arraystretch}{0.9}
		\setlength{\tabcolsep}{2.5pt}
		\resizebox{\linewidth}{!}{
			\begin{tabular}{lcccccc}
				\toprule
				Method & F1$\uparrow$ & clD$\uparrow$ & mSAP$\uparrow$ & fp-clD$\uparrow$ & Gap$\downarrow$ & P-Ang$\downarrow$ \\
				\midrule
				crYOLO & 8.2 / 38.6 & 15.4 / 33.2 & 5.1 / 25.4 & 6.5 / 18.5 & 83.4 / 56.5 & 81.2 / 70.4 \\
				YOLOv8 & 10.5 / 42.1 & 18.2 / 38.4 & 6.5 / 29.8 & 8.4 / 21.6 & 80.5 / 51.2 & 77.8 / 66.5 \\
				\midrule
				Topaz & 7.8 / 39.4 & 15.1 / 34.1 & 4.8 / 26.2 & 6.2 / 19.1 & 84.1 / 55.4 & 81.6 / 69.2 \\
				U-Net & 15.2 / 56.8 & 25.8 / 49.6 & 11.2 / 43.2 & 13.5 / 28.5 & 70.5 / 40.2 & 65.2 / 58.4 \\
				SegFormer & 17.6 / 60.5 & 28.5 / 53.4 & 12.5 / 47.8 & 14.8 / 33.2 & 67.8 / 34.5 & 63.4 / 55.2 \\
				SAM 2 & 3.5 / 43.2 & 6.8 / 40.1 & 1.5 / 29.5 & 2.4 / 22.4 & 92.4 / 50.8 & 85.2 / 70.5 \\
				\midrule
				Deep Snake & 22.4 / 68.5 & 32.5 / 55.2 & 19.8 / 58.6 & 21.4 / 43.5 & 58.5 / 23.4 & 64.2 / 48.5 \\
				CurveGCN & 24.5 / 71.4 & 34.8 / 56.5 & 22.4 / 62.5 & 23.5 / 46.8 & 55.2 / 20.5 & 61.5 / 45.2 \\
				\midrule
				\textbf{FPicker} & \textbf{31.5 / 81.4} & \textbf{42.4 / 71.4} & \textbf{28.6 / 82.9} & \textbf{34.2 / 67.9} & \textbf{48.2 / 6.8} & \textbf{52.7 / 13.7} \\
				\bottomrule
			\end{tabular}
		}
	\end{minipage}
	\hfill
	\begin{minipage}[c]{0.34\textwidth}
		\centering
		\caption{Fine-tuning Efficiency of FPicker (DLA-34) on the Custom-EMPIAR.}
		\label{tab:label_eff}
		\setlength{\tabcolsep}{2pt}
		\renewcommand{\arraystretch}{1.1}
		\resizebox{\linewidth}{!}{
			\begin{tabular}{@{}lccccc@{}}
				\toprule
				\# real imgs & 5 & 10 & 50 & 100 & 200 \\
				\midrule
				F1$\uparrow$     & 45.3 & 56.1 & 69.8 & 75.4 & 79.2 \\
				clD$\uparrow$    & 51.4 & 58.6 & 64.2 & 67.8 & 69.5 \\
				mSAP$\uparrow$   & 41.3 & 55.4 & 70.1 & 76.5 & 80.4 \\
				fp-clD$\uparrow$ & 46.2 & 54.3 & 60.5 & 63.6 & 65.8 \\
				Gap$\downarrow$  & 32.2 & 22.3 & 14.5 & 10.8 & 8.5  \\
				P-Ang$\downarrow$& 42.1 & 31.3 & 21.4 & 18.2 & 15.6 \\
				\bottomrule
			\end{tabular}
		}
	\end{minipage}
\end{table}

The substantial performance drop across all models in the zero-shot phase highlights the severe noise distribution of real micrographs. Yet, FPicker maintains the highest topological resilience. In the zero-shot regime, FPicker achieves a clDice of 42.4\% and an mSAP of 28.6\%. While the zero-shot precision reflects the immense domain gap, FPicker heavily outperforms the fine-tuned versions of established detectors (e.g., crYOLO reaches 25.4\% and Topaz achieves 26.2\% mSAP) in recall and topology conservation. Meanwhile, texture-reliant foundation models like SAM 2 experience severe zero-shot degradation (1.5\% mSAP).

Upon fine-tuning, FPicker reaches a state-of-the-art mSAP of \textbf{82.9\%}, decisively surpassing other fine-tuned baselines. This robust sim-to-real bridging indicates that our open-curve evolution successfully learns the noise-invariant physical geometry of filaments, rather than overfitting to synthetic artifacts, allowing it to seamlessly adapt its visual filters to real-world situations.

\subsection{Data Efficiency and the Sim-to-Real Paradigm}
\label{sec:sim_to_real_paradigm}
To further investigate the impact of limited annotations in Cryo-EM, we evaluate the data efficiency of our fine-tuning mechanism. As quantified in Table~\ref{tab:label_eff}, FPicker exhibits a rapid performance gain even with sparse training data: fine-tuning with just 10 images nearly doubles the zero-shot mSAP ($28.6 \rightarrow 55.4$), and expanding to 50 images yields an mSAP of $70.1$. This robust few-shot response validates its baseline feasibility for actual laboratory usage.

Fundamentally, this data-efficient adaptation suggests that the network essentially aligns with the underlying physical manifold rather than memorizing domain-specific noise. While training from scratch under extreme noise frequently leads to optimization collapse, optimizing within a physical forward model (Cryo-Sim) establishes a stable geometric prior. Consequently, on real-world EMPIAR data, the network primarily adapts its low-level visual filters rather than relearning filament definitions from sparse annotations. This observation points toward a promising Sim-to-Real paradigm for biological imaging: explicitly modeling physical priors can help mitigate the reliance on expensive manual annotations. 

Nevertheless, FPicker currently serves as a preliminary proof of concept. Fully bridging the gap between synthetic physics and real-world structural discovery remains an open challenge, requiring future explorations into more complex graph and curve topologies to handle broader geometric variations.

\subsection{Ablation Studies}
\label{sec:ablation}
To verify the individual contribution of each core component in FPicker, we conduct comprehensive ablation studies, as detailed in Table~\ref{tab:ablation}.

\textbf{Center-Endpoint Representation.}
Baseline Model A (box prior, closed curves) collapses under Extreme noise (-20 dB) with 2.5\% mSAP and a 92.5\% gap rate. Using our center-endpoint (C-E) representation (Model B) surges EMPIAR mSAP to 52.5\% and reduces the Extreme SNR gap rate to 65.4\%. This confirms C-E avoids the ghost center degeneracy by anchoring predictions directly onto the topological trajectory instead of the ambiguous background.
\begin{table}[h]
	\caption{\textbf{Component ablation across Cryo-Sim and Custom-EMPIAR.} We progressively validate Initialization (Init), Topology (Topo), Bridge Curriculum, and Auxiliary loss (Aux).}
	\label{tab:ablation}
	\centering
	\setlength{\tabcolsep}{1.1pt}
	\resizebox{\textwidth}{!}{
		\begin{tabular}{@{}l|cccc|cccccc|cccccc|cccccc@{}}
			\toprule
			& & & & & \multicolumn{6}{c|}{\textbf{Medium SNR (0.05/-13 dB)}} & \multicolumn{6}{c|}{\textbf{Extreme SNR (0.01/-20 dB)}} & \multicolumn{6}{c}{\textbf{EMPIAR (Fine-Tuned)}} \\
			\makebox[3em][c]{Model} & \makebox[3em][c]{Init} & \makebox[3em][c]{Topo} & \makebox[3em][c]{Bridge} & \makebox[3em][c]{Aux} & 
			\makebox[3em][c]{F1$\uparrow$} & \makebox[3em][c]{clD$\uparrow$} & \makebox[3.2em][c]{mSAP$\uparrow$} & \makebox[3.2em][c]{fp-clD$\uparrow$} & \makebox[3em][c]{Gap$\downarrow$} & \makebox[3em][c]{P-Ang$\downarrow$} & 
			\makebox[3em][c]{F1$\uparrow$} & \makebox[3em][c]{clD$\uparrow$} & \makebox[3.2em][c]{mSAP$\uparrow$} & \makebox[3.2em][c]{fp-clD$\uparrow$} & \makebox[3em][c]{Gap$\downarrow$} & \makebox[3em][c]{P-Ang$\downarrow$} & 
			\makebox[3em][c]{F1$\uparrow$} & \makebox[3em][c]{clD$\uparrow$} & \makebox[3.2em][c]{mSAP$\uparrow$} & \makebox[3.2em][c]{fp-clD$\uparrow$} & \makebox[3em][c]{Gap$\downarrow$} & \makebox[3em][c]{P-Ang$\downarrow$} \\
			\midrule
			A (Base) & Box & Closed & - & - & 62.1 & 70.5 & 55.4 & 68.2 & 25.5 & 52.1 & 5.8 & 15.2 & 2.5 & 3.8 & 92.5 & 87.5 & 38.4 & 35.1 & 24.5 & 22.8 & 60.4 & 75.2 \\
			B & C-E & Closed & - & - & 73.2 & 79.8 & 68.5 & 76.8 & 18.2 & 45.2 & 18.5 & 35.4 & 12.5 & 15.2 & 65.4 & 75.1 & 58.6 & 56.4 & 52.5 & 48.4 & 34.2 & 54.8 \\
			C & C-E & \textbf{Open} & - & - & 75.8 & 82.5 & 70.2 & 79.4 & 16.5 & 19.4 & 32.4 & 40.2 & 35.6 & 18.4 & 58.5 & 60.8 & 60.1 & 60.8 & 55.2 & 52.1 & 30.4 & 42.3 \\
			D & C-E & \textbf{Open} & - & \checkmark & 84.1 & 88.2 & 82.5 & 86.2 & 10.5 & 18.5 & 52.4 & 62.5 & 45.2 & 48.5 & 35.8 & 42.1 & 68.5 & 67.2 & 71.4 & 62.5 & 22.4 & 30.5 \\
			\textbf{E (Full)} & C-E & \textbf{Open} & \checkmark & \checkmark & \textbf{89.6} & \textbf{90.7} & \textbf{91.1} & \textbf{89.6} & \textbf{6.3} & \textbf{15.5} & \textbf{80.1} & \textbf{85.2} & \textbf{76.5} & \textbf{78.5} & \textbf{9.8} & \textbf{20.5} & \textbf{81.4} & \textbf{71.4} & \textbf{82.9} & \textbf{67.9} & \textbf{6.8} & \textbf{13.7} \\
			\bottomrule
		\end{tabular}
	}
\end{table}

\textbf{Open-Curve Evolution.}
Model C introduces the open-curve evolution module. While F1-score gains are moderate, P-Ang drops precipitously (45.2$^\circ$ to 19.4$^\circ$ at Medium SNR; 54.8$^\circ$ to 42.3$^\circ$ on EMPIAR). This proves explicit open-boundary graph convolutions eliminate cyclic phantom tension, solving the shrinking degeneracy at filament endpoints.

\begin{wrapfigure}{R}{0.46\textwidth}
	\centering
	\includegraphics[width=\linewidth]{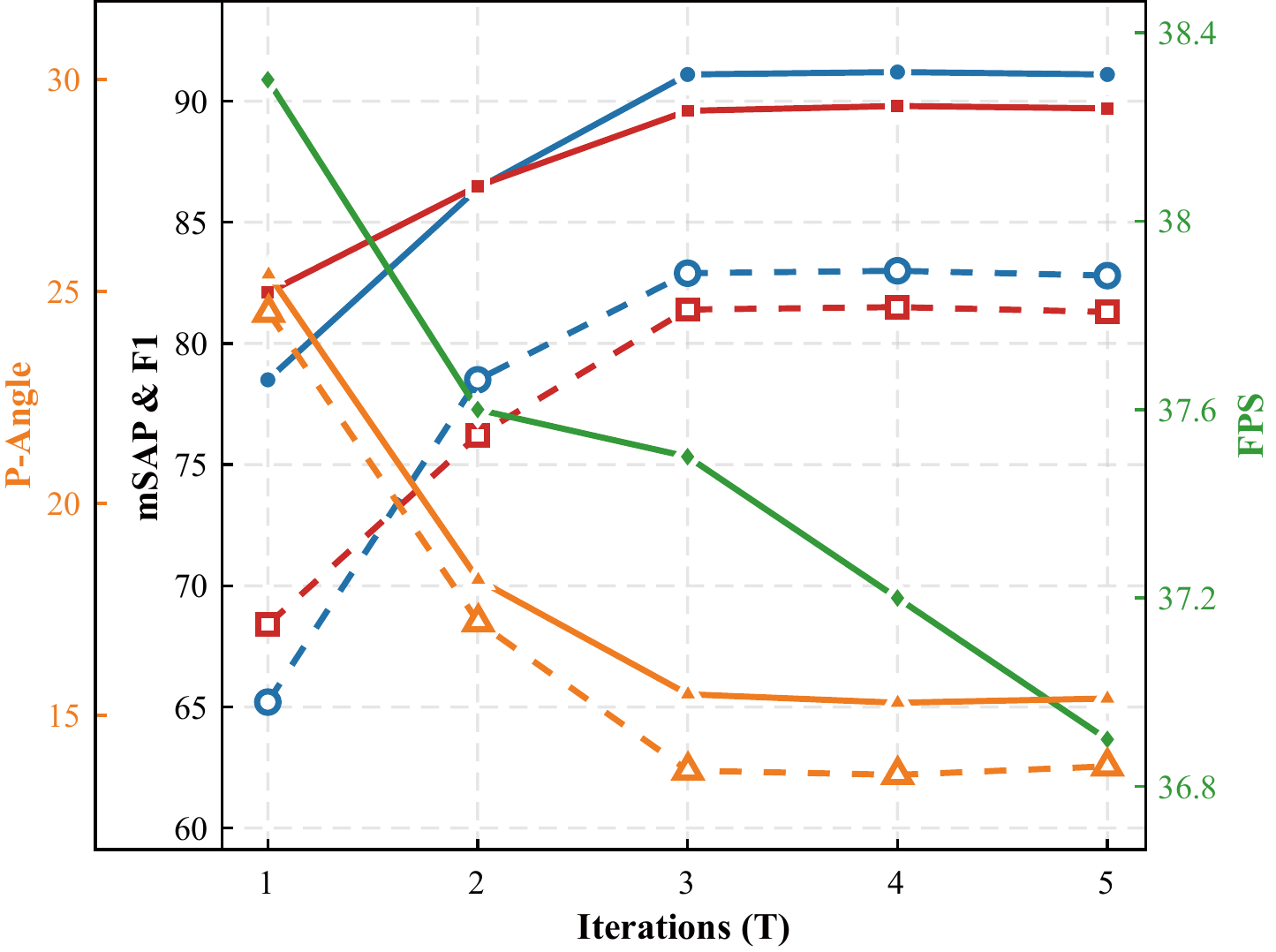}
	\caption{\textbf{Iteration Analysis.} The solid line represents Cryo-Sim, and the dashed one represents EMPIAR.}
	\label{fig:iterations}
\end{wrapfigure}

\textbf{Auxiliary Geometric Fields.}
Adding the auxiliary geometric loss (Model D) yields a transformative leap under Extreme noise: mSAP jumps from 35.6\% to 45.2\%, and gap rate nearly halves. Regressing continuous distance and angle fields forces the backbone to learn a low-pass structural filter, preventing overfitting to high-frequency Poisson noise and ensuring stable evolution guidance.

\textbf{Bridge Curriculum Strategy.}
Comparing Model D with the full FPicker (Model E), this curriculum proves vital. Early chaotic proposals otherwise trap refinement in local minima, plateauing Extreme SNR mSAP at 45.2\%. Regulating initialization variance allows Model E to reach 76.5\% mSAP at Extreme SNR and 82.9\% on EMPIAR, confirming that decoupling localization variance from topological learning is essential.

\textbf{Iteration Refinement.}
Fig.~\ref{fig:iterations} analyzes GCN deformation iterations ($T$). Performance evolves from a coarse fit at $T=1$ (78.5\% mSAP) and saturates around $T=3$ (91.1\% mSAP). Inference speed remains highly efficient (38.3 to 36.9 FPS), demonstrating that global structural anchoring achieves rapid, stable convergence without prohibitive computational overhead.

\section{Limitations}
\label{sec:limits}
FPicker precludes unified modeling of branches (\eg, Y-junctions), but true molecular branching is biologically absent in target Cryo-EM filaments. Apparent Y-junctions are merely 2D projection overlaps; resolving them as distinct filaments aligns with 3D reconstruction. Extending FPicker via dynamic node degrees for vascular or neural branching remains future work.

Furthermore, while FPicker's strong continuity prior successfully bridges noise-induced gaps, it occasionally over-merges aligned fragments (Fig.~\ref{fig:qualitative_comparison}h). Although this Over-Merging Rate (OMR, the fraction of predicted curves covering $\geq2$ GT instances) is intrinsically low (7.96\%), explicitly repelling competing tips without sacrificing noise resistance remains an open challenge.

\section{Conclusions}
\label{sec:conclus}
\textbf{FPicker} presents an effective pathway for addressing geometric incompatibilities in signal-limited regimes. By substituting rigid box priors with a center-endpoint representation and uniting holistic graph evolution with strictly open-boundary constraints, FPicker successfully overcomes local tracking derailment and cyclic shrinking, bridging the long-standing topological gap in filament tracing.

Empirically, FPicker achieves 76.5\% mSAP and a 9.8\% gap rate under Extreme noise (-20 dB), preventing ghost center collapse and endpoint shrinking. Training on Cryo-Sim cultivates a scalable geometric backbone, delivering effective zero-shot transfer and a state-of-the-art 82.9\% fine-tuned mSAP on Custom-EMPIAR. This proves modeling intrinsic physical priors outstrips brute-force data scaling in scientific imaging\cite{lecun2022path}.

\section*{Acknowledgements}
This work is supported by the National Key Research and Development Program of China under Grant 2025YFF0515300. We also acknowledge the support of the Deng Feng Fund from the School of Information Science and Technology, Tsinghua University. FPicker is a follow-up work to EPicker, developed by our group for particle extraction in Cryo-EM images. We extend our sincere gratitude to all researchers in the SGroup whose foundational efforts and prior contributions paved the way for this research.
\bibliographystyle{splncs04}
\bibliography{main}
\end{document}